\documentclass{article}

\usepackage{microtype}
\usepackage{graphicx}
\usepackage{booktabs} 

\usepackage{hyperref}

\usepackage[accepted]{icml2018}

\usepackage{amsmath}
\usepackage{amssymb}
\usepackage{tikz}
\usetikzlibrary{positioning}
\usetikzlibrary{arrows,shapes,snakes,automata,backgrounds,fit,petri,calc}
\usetikzlibrary{matrix}
\usetikzlibrary{shapes.multipart}
\usepackage{subcaption}
\usepackage{xspace}

\definecolor{Blue}{rgb}{0.0,0.0,1.0}
\definecolor{Red}{rgb}{1.0,0.0,0.0}
\definecolor{Green}{rgb}{.3,.6,0.1}

\def\drm{\mathrm{d}}

\def\Fcal{\mathcal{F}}

\def\Ncal{\mathcal{N}}

\def\Ebb{\mathbb{E}}

\def\vb{\mathbf{v}}
\def\ab{\mathbf{a}}
\def\mb{\mathbf{m}}
\def\cb{\mathbf{c}}
\def\db{\mathbf{d}}

\def\sb{\mathbf{s}}

\def\zb{\mathbf{z}}

\def\0{\mathbf{0}}

\def\vb{\mathbf{v}}

\def\xb{\mathbf{x}}

\def\Ib{\mathbf{I}}

\def\mub{\text{\boldmath $\mu$}}
\def\sigmab{\text{\boldmath $\sigma$}}

\def\varepsilonb{\text{\boldmath $\varepsilon$}}

\icmltitlerunning{Generative Temporal Models with Spatial Memory}

\begin{document}

\twocolumn[
\icmltitle{Generative Temporal Models with Spatial Memory \\for Partially Observed Environments}



\icmlsetsymbol{equal}{*}

\begin{icmlauthorlist}
\icmlauthor{Marco Fraccaro}{dtu,equal}
\icmlauthor{Danilo Jimenez Rezende}{dm}
\icmlauthor{Yori Zwols}{dm}
\icmlauthor{Alexander Pritzel}{dm}
\icmlauthor{S. M. Ali Eslami}{dm}
\icmlauthor{Fabio Viola}{dm}
\end{icmlauthorlist}

\icmlaffiliation{dtu}{Technical University of Denmark}
\icmlaffiliation{dm}{DeepMind, London, UK}

\icmlcorrespondingauthor{Marco Fraccaro}{marfraccaro@gmail.com}

\icmlkeywords{generative temporal models, spatial memory, variational inference, planning}

\vskip 0.3in
]



\printAffiliationsAndNotice{* Work done during an internship at DeepMind.} 

\begin{abstract}
In model-based reinforcement learning, generative and temporal models of environments can be leveraged to boost agent performance, either by tuning the agent's representations during training or via use as part of an explicit planning mechanism. However, their application in practice has been limited to simplistic environments, due to the difficulty of training such models in larger, potentially partially-observed and 3D environments. In this work we introduce a novel action-conditioned generative model of such challenging environments. The model features a non-parametric spatial memory system in which we store learned, disentangled representations of the environment. Low-dimensional spatial updates are computed using a state-space model that makes use of knowledge on the prior dynamics of the moving agent, and high-dimensional visual observations are modelled with a Variational Auto-Encoder. The result is a scalable architecture capable of performing coherent predictions over hundreds of time steps across a range of partially observed 2D and 3D environments.
\end{abstract}
\section{Introduction}\label{sec:intro}
Consider a setup in which an agent walks and observes an environment (e.g., a three-dimensional maze) for hundreds of time steps, and is then asked to predict subsequent observations given a sequence of actions.
This is a challenging task, as it requires the ability to first remember the visual observations and the position in which they were observed in the environment, and secondly to predict where a possibly long sequence of actions would bring the agent in the environment.
Building models that can solve this problem can be useful for model-based reinforcement learning involving spatial tasks that require long-term memories and other spatial downstream goals \citep{Sutton1990,Deisenroth2011,Levine2014,Watter2015,Wahlstrom2015,lenz2015deepmpc,Higgins2017,finn2017deep}.
This requires however agents that are able to remember the past over hundreds of steps, that know both where they are in the environment and how each action changes their position, and that can coherently predict hundreds of steps into the future.
Therefore the main focus of this work is to develop an action-conditioned generative model that is able to memorize all the required information while exploring the environment and successively use it in the prediction phase for long-term generation of high-dimensional visual observations.

Recently, several powerful generative models for sequential data have been proposed in a wide range of applications, such as modelling speech, handwriting, polyphonic music and videos \citep{Chung2015,Fraccaro2016,Oh2015,Chiappa2017}.
They build on recurrent neural architectures such as Long Short-Term Memory (LSTM) \citep{Hochreiter1997} or Gated Recurrent Units (GRU) \citep{chung2014empirical}, that use an internal state vector to perform computations and store the long-term information needed when making predictions.
Since the number of parameters in these models scales quadratically with the dimensionality of the state vector, they are not suitable for applications that require high memory capacity, such as the one considered in this paper. 
The high dimensional state vector needed to be able to memorize the hundreds of time steps in which the agent has visited the whole environment, make these models in practice both very slow and hard to train.
An alternative approach is to use an external memory architecture, for storing of large amount of information while drastically reducing the number of parameters with respect to models with similar memory capacity that build on LSTMs or GRUs.
\citet{Gemici2017} present a general architecture for generative temporal models with external memory, and test four different types of memories that are dynamically updated at each time step \citep{Graves2014,Graves2016,Santoro2016}. They focus on differentiable addressing mechanisms for memory storage and retrieval (soft-attention), 
that are based on deep neural networks that learn to write information to the memory and read from it.
While this approach is very general and can be used to model complex long-term temporal dependencies in a wide range of applications, it has not been successful in modeling the data coming from an agent freely moving in a 3d maze, even for a single room [private communications with the authors of \citep{Gemici2017}].

To define a scalable model capable of exploring larger environments and coherently predicting hundreds of time steps in the future, in this work we build a \textit{spatial memory} architecture that exploits some knowledge of the specific structure of the problem in consideration. In particular, at each time step we split the internal latent representation of the system in to two separate vectors, a low-dimensional one that encodes the position of the agent in the environment and a high dimensional one that encodes what the agent is seeing.
We model the low dimensional dynamics of the agent with a state-space model in which we encode prior information on the physical principles that govern the agent's movements, and learn a higher dimensional latent representation of the visual input (the frames from the environment) with a Variational Auto-Encoder \citep{Kingma2013,Rezende2014}. 
While exploring the environment,  at each time step we store the position of the agent and the corresponding visual information in a Differentiable Neural Dictionary (DND) \citep{Pritzel2017}, a scalable non-parametric memory developed for episodic control.
The resulting model is able to coherently generate hundreds of time steps into the future in simulated 3D environments, by retrieving at each time step the observations stored in memory that were collected when passing in nearby positions during the exploration phase. 
Making predictions with our model is scalable because of the efficient rollouts in a low dimensional space made possible by the state-space assumption and the efficient retrieval of the necessary information from DND.
The proposed model can be trained end-to-end on videos with corresponding action sequences of agents walking in an environment.
Importantly, unlike the work in \citep{Gemici2017} we do not need to learn a complex memory addressing mechanisms, as in our model the DND represents a non-parametric component where we store encodings of the positions and visual information that are learned from the data in an unsupervised way.

\section{Background}\label{sec:background}
We now provide a brief overview of the building blocks for the model introduced in section \ref{sec:theory}.

\textbf{Variational auto-encoders.} Variational auto-encoders (VAEs) \citep{Kingma2013,Rezende2014} define a generative model for high-dimensional data $\xb_t$ by introducing a latent state $\zb_t$. 
The joint probability distribution $p_\theta(\xb_t, \zb_t)$ is factorized as $p_\theta(\xb_t, \zb_t) = p_\theta(\xb_t|\zb_t)p(\zb_t)$, where $p(\zb_t)$ is the prior of the latent state and the \textit{decoder} $p_\theta(\xb_t|\zb_t)$ defines a mapping using deep neural networks parameterized by $\theta$ from the states $\zb_t$ to the data $\xb_t$. 
In a VAE, the intractable posterior distribution over the latent states is approximated using the variational distribution $q_\phi(\zb_t|\xb_t)$, also known as the \textit{encoder} or \textit{inference network}. 
The parameters $\theta$ and $\phi$ of the decoder and the encoder, respectively, are learned jointly by maximizing the Evidence Lower Bound (ELBO) with stochastic gradient ascent.

\textbf{DND memory.} The Differentiable Neural Dictionary (DND) is a scalable, non-parametric memory module first introduced in Reinforcement Learning (RL) to allow agents to store and retrieve their experiences of an environment \citep{Pritzel2017}. 
The \textit{write} operation consists of inserting (key, value) pairs into the memory; similarly to a dictionary, this associates a value to each key.
Given a query key, we can then \textit{read} from the memory by finding among the keys stored in the DND  the nearest neighbours to the query key and returning the corresponding values. 
The DND can be used in applications that require very large memories, since the nearest-neighbour search can be efficiently approximated using space-partitioning data structures, such as kd-trees \citep{Bentley1975}.

\textbf{State-space models.} State-space models (SSM) are a class of probabilistic graphical models widely used in the temporal setting to model sequences of vectors $\zb_{1:T}=[\zb_1, .., \zb_T]$ conditioned on some actions $\ab_{1:T}=[\ab_1, .., \ab_T]$.
SSMs introduce at each time step a continuous stochastic variable $\sb_t$, used as a latent representation of the state of the system.
The temporal dynamics of the system are described by the \textit{transition density} $p(\sb_{t}|\sb_{t-1}, \ab_{t})$ of the SSM, that defines how to update the state at time $t$ given the previous state $\sb_{t-1}$ and the current action $\ab_t$. 
The output variable $\zb_t$ depends on the state $\sb_t$ through the \textit{emission density} $p(\zb_t|\sb_t)$.

\section{Model}\label{sec:theory}

\begin{figure*}[t]
\centering
\def\col{blue}
\begin{subfigure}[t]{0.26\textwidth}
\centering	  
  \begin{tikzpicture}[bend angle=45,>=latex,font=\small,scale=0.9, every node/.style={transform shape}]%

	\tikzstyle{obs} = [ circle, thick, draw = black!100, fill = blue!10, minimum size = 0.8cm, inner sep = 0pt]
	\tikzstyle{lat} = [ circle, thick, draw = black!100, fill = red!0, minimum size =  0.8cm, inner sep = 0pt]
	\tikzstyle{par} = [ circle, thin, draw, fill = black!100, minimum size = 0.8, inner sep = 0pt]
	\tikzstyle{det} = [ diamond, thick, draw = black!100, fill = red!0, minimum size = 1cm, inner sep = 0pt]
	\tikzstyle{inv} = [ circle, thin, draw=white!100, fill = white!100, minimum size = 0.8cm, inner sep = 0pt]
	\tikzstyle{annotation} = [rectangle, thin, draw=white, fill=white ]
	\tikzstyle{every label} = [black!100]%
	\begin{scope}[node distance = 1.4cm and 1.4cm, rounded corners=4pt]

	    \node (z_tm2) [] {};
		\node [lat] (z_tm1) [ right of = z_tm2]   {$\sb_{t-1}$};
		\node [lat] (z_t) [ right of = z_tm1] {$\sb_{t}$};
		\node (z_tp1) [ right of = z_t]{};

        \draw[post] (z_tm2) edge  (z_tm1);
		\draw[post] (z_tm1) edge  (z_t);
		\draw[post] (z_t) edge  (z_tp1);

		\node [rectangle, draw=red] (m_tm1) [ below = 0.15cm of z_tm1]   {Memory};
		\node [rectangle, draw=red] (m_t) [ below = 0.15cm of z_t] {Memory};

		\draw[-,draw=red,fill=red] (z_tm1) edge  (m_tm1);
		\draw[-,draw=red,fill=red] (z_t) edge  (m_t);

		\node [lat] (a_tm1) [ below = 0.35cm of m_tm1]   {$\zb_{t-1}$};
		\node [lat] (a_t) [ below = 0.35cm of m_t] {$\zb_{t}$};

		\draw[post,draw=red,fill=red] (m_tm1) edge  (a_tm1);
		\draw[post,draw=red,fill=red] (m_t) edge  (a_t);


		\node [obs] (x_tm1) [ below of = a_tm1]   {$\xb_{t-1}$};
		\node [obs] (x_t) [ below of = a_t] {$\xb_{t}$};

		\draw[post] (a_tm1) edge  (x_tm1);
		\draw[post] (a_t) edge  (x_t);

		\node [obs] (u_tm1) [ above of = z_tm1]   {$\ab_{t-1}$};
		\node [obs] (u_t) [ above of = z_t] {$\ab_{t}$};

		\draw[post] (u_tm1) edge  (z_tm1);
		\draw[post] (u_t) edge  (z_t);

		\draw[post] (x_tm1) edge[bend left, dashed]  (a_tm1);
		\draw[post] (x_t) edge[bend left, dashed]  (a_t);

		\node[annotation] (vae) at (3.65cm, -3.0cm) {VAE};

		\node[annotation] (lgssm) at (3.8cm, 0.6cm) {SSM};

		\begin{pgfonlayer}{background}
	    \filldraw [dashed, line width = 1pt, draw=blue!40, fill=white]
		($(a_tm1.south west) + (-0.7cm, -0.3cm)$) rectangle ($(u_t.north east) + (0.7cm, 0.3cm)$);
		\filldraw [dashed, line width = 1pt, draw=black!40!green, fill=white]
		($(x_t.south west) + (-0.3cm, -0.3cm)$) rectangle ($(a_t.north east) + (0.3cm, 0.3cm)$);
     	\end{pgfonlayer}
	\end{scope}%
\end{tikzpicture}%
\caption{Generative model of the GTM-SM. Red arrows represent dependencies on the DND memory. Dashed arrows represent the VAE inference network.}%
\label{fig:ssmm_gen}%
\end{subfigure}
\hfill
\begin{subfigure}[t]{0.42\textwidth}
\centering	  
\begin{tikzpicture}[>=latex]
  \tikzstyle{lat} = [ circle, thick, draw = black!100, fill = red!0, minimum size =  0.8cm, inner sep = 0pt]

  \node [lat] (z_t) {$\sb_{t}$};

  \node[rectangle split, rectangle split parts=10, 
       draw, minimum width=0.8cm,font=\small,
       rectangle split part align={center}] (t1) [ right = 0.8cm of z_t]
    {             $\sb_1$
     \nodepart{two}
                  $\sb_2$
     \nodepart{three}
                  $\sb_3$  
     \nodepart{four}
                  $\sb_4$ 
     \nodepart{five}
                  $\sb_5$ 
     \nodepart{six}
                  $\sb_6$ 
     \nodepart{seven}
                  $\sb_7$ 
     \nodepart{eight}
                  $\sb_8$
     \nodepart{nine}
                  $\dots$ 
     \nodepart{ten}
                  $\sb_{256}$};
                  
  \node [] (m) [ above right = 0cm and -0.3cm of t1] {$\mb$};

  \draw[->, red] (z_t) to [out=0, in=180](t1.text west);      
  \draw[->, red] (z_t) to [out=0, in=180](t1.two west);      
  \draw[->, red] (z_t) to [out=0, in=180](t1.three west);      
  \draw[->, red] (z_t) to [out=0, in=180](t1.four west);      
  \draw[->, red] (z_t) to [out=0, in=180](t1.five west);      
  \draw[->, red] (z_t) to [out=0, in=180](t1.six west);      
  \draw[->, red] (z_t) to [out=0, in=180](t1.seven west);      
  \draw[->, red] (z_t) to [out=0, in=180](t1.eight west);      
  \draw[->, red] (z_t) to [out=0, in=180](t1.nine west);      
  \draw[->, red] (z_t) to [out=0, in=180](t1.ten west);        

  \node[rectangle split, rectangle split parts=10, 
       draw, minimum width=0.8cm,font=\small,
       rectangle split part align={center}] (t2) [ right = 0cm of t1]
    {             $\zb_1$
     \nodepart{two}
                  $\zb_2$
     \nodepart{three}
                  $\zb_3$  
     \nodepart{four}
                  $\zb_4$ 
     \nodepart{five}
                  $\zb_5$ 
     \nodepart{six}
                  $\zb_6$ 
     \nodepart{seven}
                  $\zb_7$ 
     \nodepart{eight}
                  $\zb_8$
     \nodepart{nine}
                  $\dots$ 
     \nodepart{ten}
                  $\zb_{256}$};

  \node[rectangle split, rectangle split parts=3, 
       draw, minimum width=0.4cm, font=\small,
       rectangle split part align={center}] (t3) [ right = 0.4cm of t2]
    {             \scalebox{.69}{$d_2$}
     \nodepart{two}
                  \scalebox{.69}{$d_6$}
     \nodepart{three}
                  \scalebox{.69}{$d_{125}$}};

  \node[rectangle split, rectangle split parts=3, 
       draw, minimum width=0.8cm,font=\small,
       rectangle split part align={center}] (t4) [ right = 0cm of t3]
    {             $\sb_2$
     \nodepart{two}
                  $\sb_6$
     \nodepart{three}
                  $\sb_{125}$};
                  
  \node[rectangle split, rectangle split parts=3, 
       draw, minimum width=0.8cm,font=\small,
       rectangle split part align={center}] (t5) [ right = 0cm of t4]
    {             $\zb_2$
     \nodepart{two}
                  $\zb_6$
     \nodepart{three}
                  $\zb_{125}$};

  \draw[->, red] (t2.two east) to [out=0, in=180](t3.text west);      
  \draw[->, red] (t2.six east) to [out=0, in=180](t3.two west);      
  \draw[->, red] (t2.nine east) to [out=0, in=180](t3.three west);      

  \node [lat] (a_t) [ right = 0.4cm of t5] {$\zb_{t}$};

  \draw[->, red] (t5.text east) to [out=0, in=174](a_t);      
  \draw[->, red] (t5.two east) to [out=0, in=177](a_t);      
  \draw[->, red] (t5.three east) to [out=0, in=183](a_t);   

		\begin{pgfonlayer}{background}
	    \filldraw [line width = 0.5pt, draw=red!70, fill=white, rounded corners=.4cm]
		($(a_tm1.south west) + (-0.45cm, -0.25cm)$) rectangle ($(u_t.north east) + (3.05cm, 0.95cm)$);
     	\end{pgfonlayer}
     	
     	\node[rectangle, thin, draw=white, fill=white] (dnd) at (3.4cm, 2.8cm) {DND memory};

\end{tikzpicture}
\caption{The dependence of $\zb_t$ on $\sb_t$ and the DND memory $\mb$ in the VAE prior $p(\zb_t|\sb_t,\mb)$. This corresponds to the red rectangle in Figure \ref{fig:ssmm_gen}. Here, the memory contains the first $\tau=256$ time steps of the video, and we retrieve the $K=3$ closest neighbours to $\sb_t$ ($d$ is the distance).}%
\label{fig:ssmm_dnd}%
\end{subfigure}
\hfill
\begin{subfigure}[t]{0.26\textwidth}
\centering	  
  \begin{tikzpicture}[bend angle=45,>=latex,font=\small,scale=0.9, every node/.style={transform shape}]%

	\tikzstyle{obs} = [ circle, thick, draw = black!100, fill = blue!10, minimum size = 0.8cm, inner sep = 0pt]
	\tikzstyle{lat} = [ circle, thick, draw = black!100, fill = red!0, minimum size =  0.8cm, inner sep = 0pt]
	\tikzstyle{par} = [ circle, thin, draw, fill = black!100, minimum size = 0.8, inner sep = 0pt]
	\tikzstyle{det} = [ diamond, thick, draw = black!100, fill = red!0, minimum size = 1cm, inner sep = 0pt]
	\tikzstyle{inv} = [ circle, thin, draw=white!100, fill = white!100, minimum size = 0.8cm, inner sep = 0pt]
	\tikzstyle{annotation} = [rectangle, thin, draw=white, fill=white ]
	\tikzstyle{every label} = [black!100]%
	\begin{scope}[node distance = 1.4cm and 1.4cm, rounded corners=4pt]

	    \node (z_tm2) [] {};
		\node [lat] (z_tm1) [ right of = z_tm2]   {$\sb_{t-1}$};
		\node [lat] (z_t) [ right of = z_tm1] {$\sb_{t}$};
		\node (z_tp1) [ right of = z_t]{};

        \draw[post] (z_tm2) edge  (z_tm1);
		\draw[post] (z_tm1) edge  (z_t);
		\draw[post] (z_t) edge  (z_tp1);

		\node [lat] (a_tm1) [ below = 0.15cm of m_tm1]   {$\zb_{t-1}$};
		\node [lat] (a_t) [ below = 0.15cm of m_t] {$\zb_{t}$};


		\node [obs] (x_tm1) [ below of = a_tm1]   {$\xb_{t-1}$};
		\node [obs] (x_t) [ below of = a_t] {$\xb_{t}$};

		\node [obs] (u_tm1) [ above of = z_tm1]   {$\ab_{t-1}$};
		\node [obs] (u_t) [ above of = z_t] {$\ab_{t}$};

		\draw[post] (u_tm1) edge  (z_tm1);
		\draw[post] (u_t) edge  (z_t);

		\draw[post] (x_tm1) edge[bend left]  (a_tm1);
		\draw[post] (x_t) edge[bend left]  (a_t);

	\end{scope}%
\end{tikzpicture}%
\caption{Inference network for the GTM-SM.}%
\label{fig:ssmm_inf}%
\end{subfigure}
    \caption{Generative Temporal Model with Spatial Memory}\label{fig:ssmm}
\end{figure*}
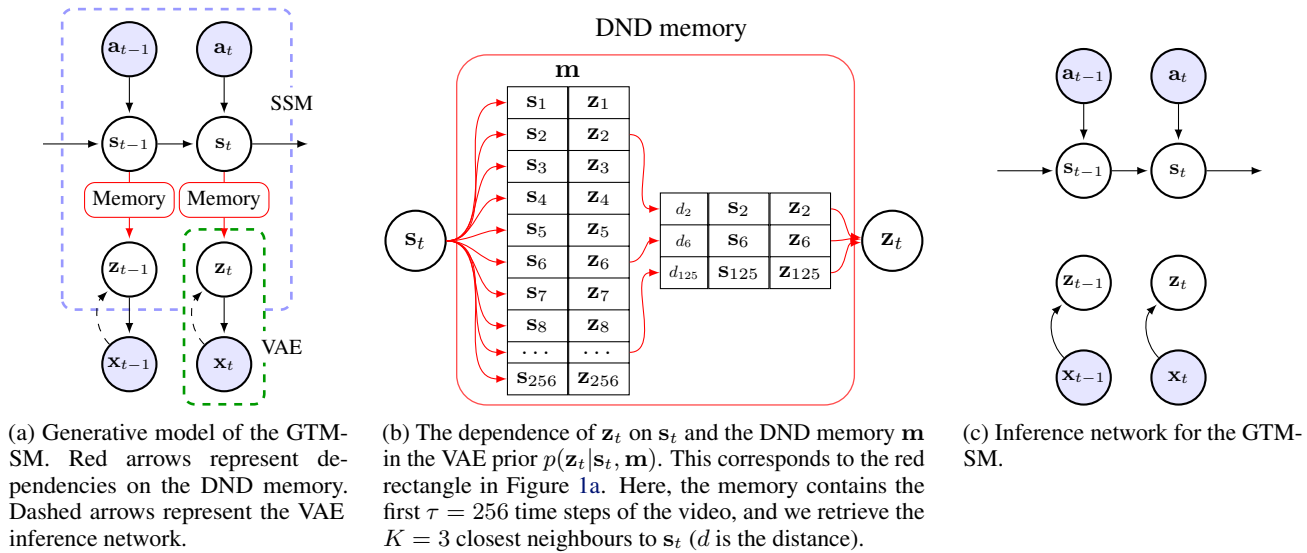

An important component of model-based reinforcement learning is the ability to plan many steps ahead in time leveraging previous experiences \citep{Sutton1990,Racaniere2017}. 
This requires agents that can remember the past and use it to predict what may happen in the future given certain actions. 
With this purpose in mind, we define an action-conditioned generative model with memory, that can be used within RL agents for model-based planning.

The input of our model consists of $T$-step videos with corresponding action sequences, generated by an agent acting in an environment. We split each sequence of $T$ time steps into two parts, corresponding to two different model phases:
\begin{enumerate}
\item \textit{Memorization phase}. For $t=1,..,\tau$, the model receives at each time step a frame $\xb_{t}$ and action $\ab_{t}$ (e.g. move forwards/backwards, rotate left/right) that led to it. In this phase, the model has to store in memory all the information needed in the following prediction phase. 
During this phase the agent sees most of the environment (but from a restricted set of viewpoints), in order to give the model sufficient information to make accurate predictions in the subsequent phase.

\item \textit{Prediction phase}. For $t=\tau+1,..,T$, the model receives the actions $\ab_{\tau+1:T}$ that move the data-generating agent across the previously explored environment (although perhaps viewed from a different angle) and needs to predict the observations $\xb_{\tau+1:T}$ using the information about the past that is stored in the memory.
\end{enumerate}

Storing \textit{what} the agent sees at each time step is not sufficient: in order to retrieve the correct information from memory when returning to the same location during the prediction phase, we also need to store \textit{where} the agent is.
The location of the agent is a latent variable that can be inferred given the actions as explained in the rest of this section.

As shown in Figure \ref{fig:ssmm_gen}, in a \textit{Generative Temporal Model with Spatial Memory (GTM-SM)} 
 we introduce two sets of latent variables that disentangle visual and dynamics information, similarly to \citep{Fraccaro2017}. 
At each time step we have a VAE whose latent state $\zb_t$ is an encoding of the frame of the video and therefore captures the visual information. 
The priors of the VAEs are temporally dependent through their dependence on the states $\sb_t$ of the SSM, a latent representation of the location of the agent in the environment. 
The transition density of the SSM is used to include prior knowledge on the environment dynamics, i.e. the underlying physics.

During the initial memorization phase, the GTM-SM infers the states $\sb_{1:\tau}$ of the agent (i.e. the position) and the frame encodings $\zb_{1:\tau}$, and stores these $(\sb_i, \zb_i)$ pairs as (key, value) in the DND memory. Probabilistically, we can view this as inserting an approximation of the intractable the posterior $p(\sb_{1:\tau}, \zb_{1:\tau}|\xb_{1:\tau}, \ab_{1:\tau})$ into the DND; see Section \ref{sec:past_enc} for details.
As the latent variables are stochastic, in practice we store in memory the sufficient statistics of the distribution (e.g. the mean and variance in the case of Gaussian variables). To keep the notation simple, we will refer to the information in the DND by the name of the random variable (as done in Figure \ref{fig:ssmm_dnd}), rather than introducing a new symbol for the sufficient statistics.
An alternative is to insert one or more samples from the distribution into the memory instead, but this would introduce some sampling noise.

In the subsequent prediction phase, we forward-generate from the SSM using the actions $\ab_{\tau+1:T}$ to predict $\sb_{\tau+1:T}$, and we use the VAE's generative model to generate the frames $\xb_{\tau+1:T}$ given the predicted states and the information from the first $\tau$ time steps stored in the DND memory; see Section \ref{sec:gen} for details.
In our experiments, a low-dimensional state vector $\sb_t$  (2- or 3-dimensional) suffices. Because of this, we can perform efficient rollouts in latent space without the need to generate high-dimensional frames at each time step as in autoregressive models \citep{Oh2015,Chiappa2017,Gemici2017}.
Also, thanks to the scalability properties of the DND memory, we can efficiently explore very large environments.

There are three key components that define the GTM-SM and that will be introduced in the following, namely the \textit{generative model}, the \textit{inference network}, and the \textit{past encoder}. As we will see, these components share many parameters.

\subsection{Generative model}\label{sec:gen}
For brevity, we write the observations and actions in the memorization phase as $\vb=\{\xb_{1:\tau}, \ab_{1:\tau} \}$ and we write the observations and actions in the prediction phase as $\xb=\xb_{\tau+1:T}$ and $\ab=\ab_{\tau+1:T}$ respectively.
Letting $\theta$ be the parameters of the generative model, we model $p_\theta(\xb| \ab, \vb)$ as follows.
We introduce two sets of latent variables: the frame encodings $\zb=\zb_{\tau+1:T}$ and the SSM states $\sb=\sb_{\tau+1:T}$, and define the joint probability density $p_\theta(\xb, \zb, \sb| \ab, \vb)$ following the factorization shown in Figure \ref{fig:ssmm_gen}:
\begin{align}
p_\theta(\xb, \zb, \sb| \ab, \vb) & =\prod_{t=\tau+1}^T p_\theta(\xb_t|\zb_t)p_\theta(\zb_t|\sb_t,\mb) \cdot \nonumber \\
& \qquad \qquad \cdot p_\theta(\sb_{t}|\sb_{t-1}, \ab_{t}),
\end{align}
where $p_\theta(\sb_{t}|\sb_{t-1}, \ab_{t})$ is a Gaussian SSM transition probability density and $p_\theta(\xb_t|\zb_t)$ is the VAE decoder (Bernoulli or Gaussian distributed, depending on the data). 
$p_\theta(\zb_t|\sb_t,\mb)$ can be seen as the prior of the VAE, that is conditioned at each time step on the current state $\sb_t$ and the content of the DND memory $\mb=\{\sb_{1:\tau}, \zb_{1:\tau} \}=\mathit{PastEncoder}(\vb)$ as illustrated in Figure \ref{fig:ssmm_dnd} (see Section \ref{sec:past_enc} for details on the past encoder). Its sufficient statistics are computed as follows.
First, we calculate the distances $d_i=d(\sb_i,\sb_t)$, $i=1, ..,\tau$, between $\sb_t$ and all the states $\sb_i$ in the DND memory. 
We then retrieve from the memory the $K$ nearest states and the corresponding frame encodings, thus forming a set of triplets $\{(d^{(k)}, \sb^{(k)},\zb^{(k)}), k=1, .. ,K\}$ that will be used as conditioning variables when computing the parameters of the VAE prior $p_\theta(\zb_t|\sb_t,\mb)$.
Using low-dimensional $\sb_t$ and prior knowledge of the environment dynamics when defining $p_\theta(\sb_{t}|\sb_{t-1}, \ab_{t})$, we can make the GTM-SM learn to use $\sb_t$ to represent its position in the environment. 
At each time step the model will then retrieve from the memory what it has seen when it was previously close to the same location, and use this information to generate the current frame $\xb_t$.
The exact form of the VAE prior $p_\theta(\zb_t|\sb_t,\mb)$ and transition model $p_\theta(\sb_{t}|\sb_{t-1}, \ab_{t})$ is environment-dependent, and will be therefore introduced separately for each experiment in Section \ref{sec:experiments}.

\subsection{Inference network}\label{sec:inf}
Due to the non-linearities in the VAE and the fact that $p_\theta(\zb_t|\sb_t,\mb)$ depends on the DND memory, the posterior distribution $p_\theta(\zb, \sb| \xb, \ab, \vb)$ of the GTM-SM is intractable.
We therefore introduce a variational approximation $q_\phi(\zb, \sb| \xb, \ab)$ that factorizes as
\begin{align}
q_\phi(\zb, \sb| \xb, \ab) & = q_\phi(\zb| \xb) p_\theta(\sb|\ab)\nonumber\\
& = \prod_{t=\tau+1}^T q_\phi(\zb_t| \xb_t) p_\theta(\sb_t| \sb_{t-1}, \ab_t) \ .\label{eq:infnet}
\end{align}
A graphical representation of the inference network of the GTM-SM is shown in Figure \ref{fig:ssmm_inf}.
$q_\phi(\zb_t| \xb_t)$ is an inference network that outputs the mean and variance of a Gaussian distribution, as typically done in VAEs. 
In \eqref{eq:infnet} we then use the SSM transition probabilities, and we are therefore assuming that the GTM-SM can learn the prior dynamics of the moving agents accurately enough to infer the position of the agent given the sequence of actions.
Notice that without this assumption it would be impossible to perform long term generation with the model during the prediction phase. In this phase, we can in fact only rely on the generative model, and not on the inference network as we do not know what the agent is seeing at each time step.
To relax this assumption, the inference network could be extended to make use of the information stored in memory, for example by using landmark information when inferring the current position of the agent. 
This is discussed more in detail in Appendix \ref{sec:inf_landmark} in the supplementary material, together with an initial experiment to assess the feasibility of the proposed method.

\subsection{Past encoder}\label{sec:past_enc}
The past encoder is used during the memorization phase to extract the information to store in the DND memory. 
It creates a mapping from $\sb_t$ to $\zb_t$ that is exploited at times $t=\tau+1:T$ in the generative model.
During the memorization phase, at each time step we store in the DND the sufficient statistics of the inferred states $\sb_t$ and visual information $\zb_t$,  $t=1,.., \tau$, obtained from an approximation to the smoothed posterior $p_\theta(\zb_t, \sb_t|\vb)$. We first factorize this distribution as
$
p_\theta(\zb_t, \sb_t|\vb)=p_\theta(\sb_t|\vb) p_\theta(\zb_t|\vb)
$
and only condition on the information up to time $t$ (i.e. we are doing \textit{filtering} instead of smoothing):
$
p_\theta(\zb_t,\sb_t|\vb) \approx p_\theta(\sb_t|\vb_{1:t}) p_\theta(\zb_t|\vb_{1:t})  
$, where $\vb_{1:t}=\{\xb_{1:t}, \ab_{1:t} \}$
Second, we approximate each of the terms using the inference network, without introducing any additional parameters in the model.
Using the VAE encoder, we have $p_\theta(\zb_t|\vb_{1:t}) \approx q_\phi(\zb_t|\xb_t)$.
We then assume $p_\theta(\sb_t|\vb_{1:t}) \approx p_\theta(\sb_t|\vb_{1:t-1}, \ab_t) $ with
\begin{align*}
 p_\theta(\sb_t|\vb_{1:t-1}, \ab_t) 
 = \int p_\theta(\sb_t| \sb_{t-1}, \ab_t)p_\theta^\star(\sb_{t-1}) \ \drm \sb_{t-1}\ ,
\end{align*}
\begin{align}\label{eq:marg_s}
p_\theta^\star(\sb_{t}) = \int p_\theta(\sb_{t}|\sb_{t-1},\ab_{t}) p_\theta^\star(\sb_{t-1}) \ \drm \sb_{t-1} \ .
\end{align}
$p_\theta^\star(\sb_{t})$ is the marginal distribution of $\sb_{t}$ obtained by integrating over the past states (samples from $p_\theta^\star(\sb_{t})$, as needed in the ELBO, are easily obtained with ancestral sampling).

\subsection{Training}\label{sec:elbo}
We learn the parameters $\theta$ and $\phi$ of the GTM-SM by maximizing the ELBO, a lower bound to the log-likelihood $\log p_\theta(\xb| \ab, \vb)$ obtained using Jensen's inequality and the inference network introduced in Section \ref{sec:inf}:
\begin{align*}
\log  p_\theta&(\xb| \ab, \vb)=\log \int p_\theta(\xb, \zb, \sb| \ab, \vb) \ \drm \zb \ \drm \sb \nonumber \\
&\geq\Ebb_{q_\phi(\zb, \sb| \xb, \ab, \vb)} \left[ \log \frac{p_\theta(\xb, \zb, \sb| \ab, \vb)}{q_\phi(\zb, \sb| \xb, \ab, \vb)} \right] = \Fcal(\theta , \phi)   \ .
\end{align*}
Exploiting the temporal factorization of both the joint distribution $p_\theta(\xb, \zb, \sb| \ab, \vb)$ and the variational approximation $q_\phi(\zb, \sb| \xb, \ab, \vb)$, we obtain after some calculations:
\begin{align*}
 \Fcal(\theta , \phi) & =\sum_{t=\tau+1}^T \Ebb_{q_\phi(\zb_t|\xb_t)} \left[ \log p_\theta(\xb_t|\zb_t) \right] +\\ 
 & \qquad \quad - \Ebb_{p_\theta^\star(\sb_{t})} \left[ \mathit{KL} [q_\phi(\zb_t|\xb_t) ||  p_\theta(\zb_t|\sb_t,\mb) ]  \right] \ .
\end{align*}
The ELBO is then formed by two terms: a reconstruction term and an expected KL divergence for the VAE, which depends on the transition model through the expectation over the marginal distribution $p_\theta^\star(\sb_{t})$ in \eqref{eq:marg_s}. 
$\Fcal(\theta , \phi)$ can be maximized jointly with respect to $\theta$ and $\phi$ with stochastic gradient ascent, approximating the intractable expectations with Monte Carlo integration with a single sample and using the reparameterization trick to obtain low-variance gradients \citep{Kingma2013,Rezende2014}.
For a given training sequence, we compute the distribution needed in the ELBO as follows: first, we use the past encoder to extract the information to be stored in the DND memory $\mb$. 
We then obtain samples from the distribution $p_\theta^\star(\sb_{t})$ by iteratively applying the transition model, and use them to read from the DND memory as needed to compute the VAE prior $p_\theta(\zb_t|\sb_t,\mb)$. 
Finally, the variational approximation and the likelihood are computed as usual in VAEs.
\section{Experiments}\label{sec:experiments}

We test the memorization and long-term generation capabilities of the GTM-SM on several 2D and 3D environments of increasing complexity.
We use videos with action data from RL agents walking in the environments, that are split so that both the memorization and prediction phase are hundreds of time steps.
Experimental details can be found in Appendix \ref{sec:exp_details}. Videos of long-term generations from the model for all the experiments are available in the supplementary material, see Appendix \ref{sec:app_video} for details.

\subsection{Image navigation experiment}\label{sec:imagenav}

\begin{figure*}[t]
\centering
\def\col{blue}
\begin{subfigure}[t]{0.32\textwidth}
\centering	  
\includegraphics[width=0.65\textwidth]{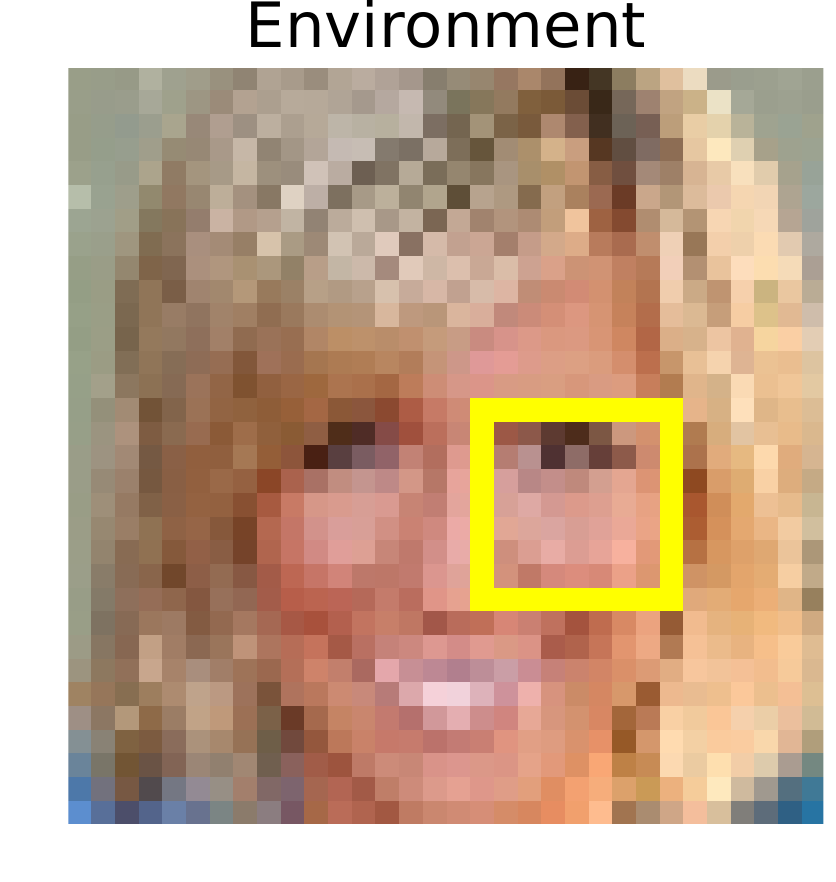}
\caption{Environment and current observation (the yellow square).}%
\label{fig:image_nav_states_env}%
\end{subfigure}
\hfill
\begin{subfigure}[t]{0.32\textwidth}
\centering	  
\includegraphics[width=0.7\textwidth]{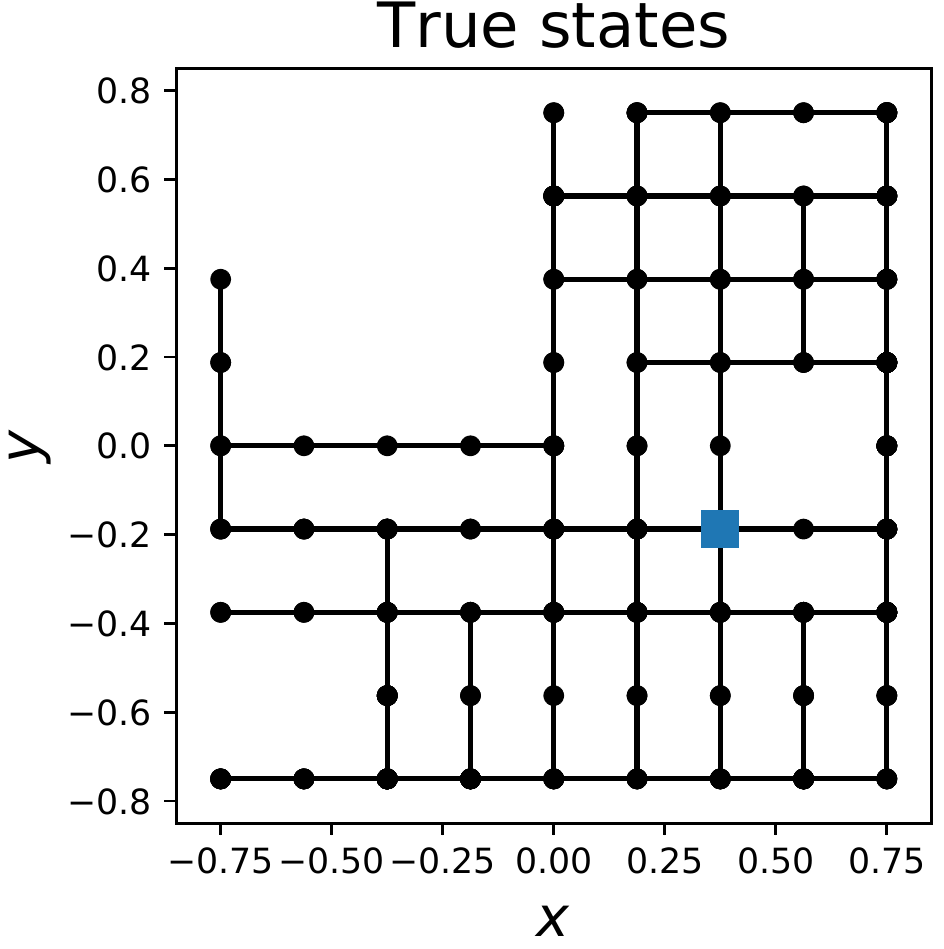}
\caption{Ground truth positions explored during the 256 steps of the memorization phase. The blue square represents the final position.}%
\label{fig:image_nav_states_true}%
\end{subfigure}
\hfill
\begin{subfigure}[t]{0.32\textwidth}
\centering	  
\includegraphics[width=0.75\textwidth]{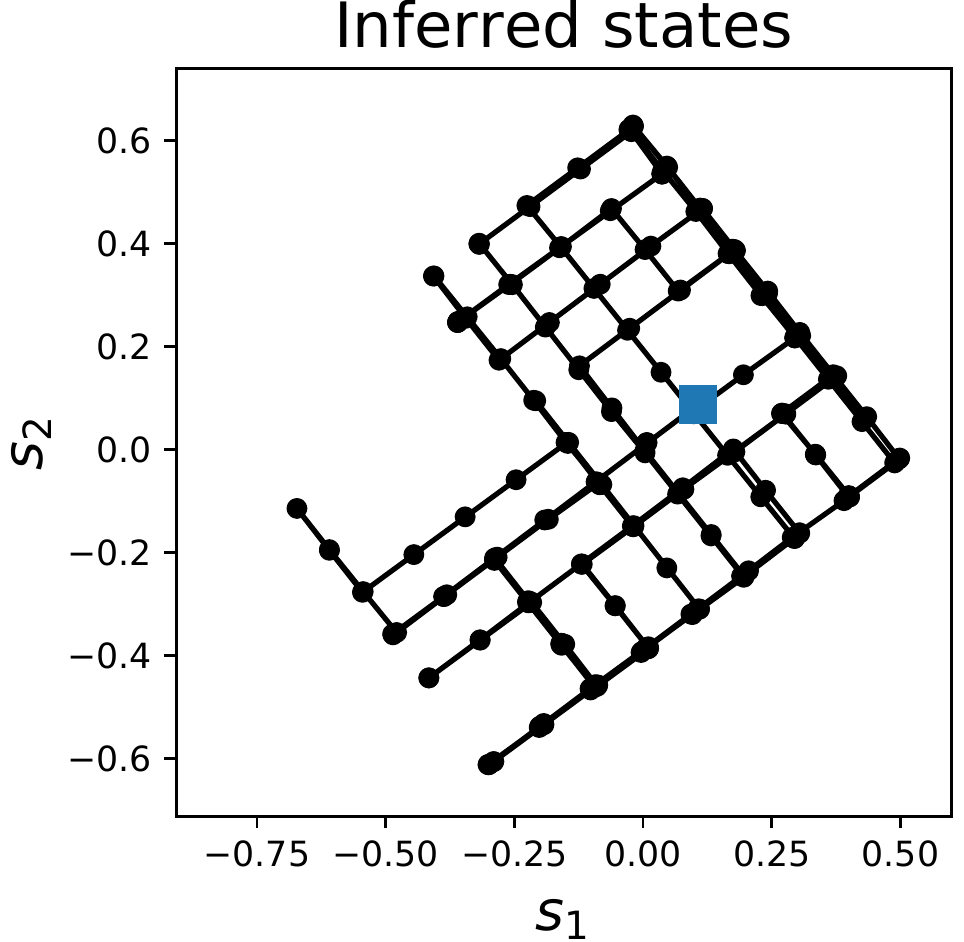}
\caption{Inferred positions during the 256 steps of the memorization phase, $\sb_t=(s_1, s_2)$.}%
\label{fig:image_nav_states_infer}%
\end{subfigure}

\vspace*{0.2cm}

\begin{subfigure}[t]{\textwidth}
\centering	  
\includegraphics[width=0.97\textwidth]{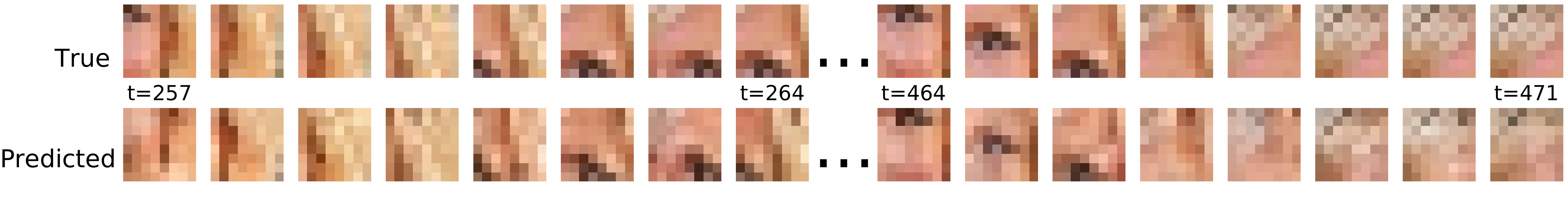}
\caption{Comparison between true and predicted frames during the prediction phase of a test sequence (time steps $t_{257:264}$ and $t_{464:471}$).}%
\label{fig:image_nav_gen}%
\end{subfigure}
\caption{Image navigation experiment}\label{fig:image_nav}
\end{figure*}

In this experiment the data-generating agent walks on top of an image and observes a cropped version of the image (centered at the agent's position).
As illustrated in Figure \ref{fig:image_nav_states_env}, the 2D environment is a 32x32 image from the CelebA dataset \citep{Liu2015} and the agent sees an 8x8 crop (the yellow square in the figure). 
There are five possible actions: move one step up/down/left/right or stay still.
At each time step we sample a random action, but to favor exploring the whole environment the action is repeated in the subsequent time steps. The number of repetitions is sampled from a Poisson distribution. 
The agent cannot walk outside of the image: the ``move right'' action, for example, does not change the position of an agent on the right edge of the image. 
We can interpret this as an environment with walls.
The agent walks on the image while adding information in the DND memory for $\tau=256$ time steps. During training the prediction phase has 32 time steps; during testing we instead generate from the model for 256 time steps, so that $T=512$. 
In each of the two dimensions, there are nine possible positions (the crops can overlap). This is illustrated in Figure \ref{fig:image_nav_states_true}, which shows the ground truth positions that the agent has visited in the 256 steps of the memorization phase of a test sequence.

We use a 2-dimensional state space that the GTM-SM learns to use to represent the position of the agent.
With no walls in the environment all possible transitions are linear, and they can be modelled as 
$\sb_t = \sb_{t-1} + M\ab_t + \varepsilonb_t$ with $\varepsilonb_t \sim \Ncal(\mathbf{0}, r^2\Ib)$. In all experiments we use small values for $r^2$, that make the transitions close to being deterministic. The transition matrix $M$ can be learned from the data, and describes how to update the state given each of the 5 actions ($M\ab_t$ is the \textit{displacement} at time $t$). 
The environment in our experiment has walls however, and we need the model to be able to learn not to move with actions that would make it hit a wall.
We do this by multiplying the displacement by a neural network $\sigma_d$ that receives as input the projected position of the agent after taking the action and outputs a value between 0 and 1, and that can therefore learn to cancel out any displacements that would bring the agent out of the environment.
These non-linear transitions are therefore modelled as
\begin{equation}\label{eq:nonlin_tr}
\sb_t = \sb_{t-1} + M\ab_t \cdot \sigma_d (\sb_{t-1} + M\ab_t) + \varepsilonb_t \ .
\end{equation}
The VAE prior used in this experiments is obtained by creating a mixture distribution from the sufficient statistics of the frame encodings retrieved from the DND memory, see Appendix \ref{app:imagenav} for details.

In Figure \ref{fig:image_nav_states_infer} we show an example of the states inferred by the model for a test sequence. We see that the model has learned the correct transitions, in a state space that is rotated and stretched with respect to the ground truth one.
To test the memorization and prediction capabilities of the GTM-SM, Figure \ref{fig:image_nav_gen} shows a comparison between the ground truth frames of the video and the predicted ones during the prediction phase. The model produces almost perfect predictions, even after more than 200 generation steps ($t=471$). This shows that it has learned to store all relevant information in the DND, as well as retrieve all relevant information from it.
The state-of-the-art generative temporal models with memory introduced in \citep{Gemici2017}, are not able to capture the spatial structure of large environments as in the GTM-SM, and would therefore struggle to coherently generate hundreds of time steps into the future.
The MNIST maze experiment in \citep{Gemici2017} can be seen as a simpler version of the image navigation experiment presented above, with agents moving on a 4x4 grid, linear transitions and 25-step sequences.

\subsection{Labyrinth experiments}\label{sec:labyr}

\begin{figure*}[t]
\centering
\begin{subfigure}[t]{0.49\textwidth}
\centering	  
\includegraphics[width=\textwidth]{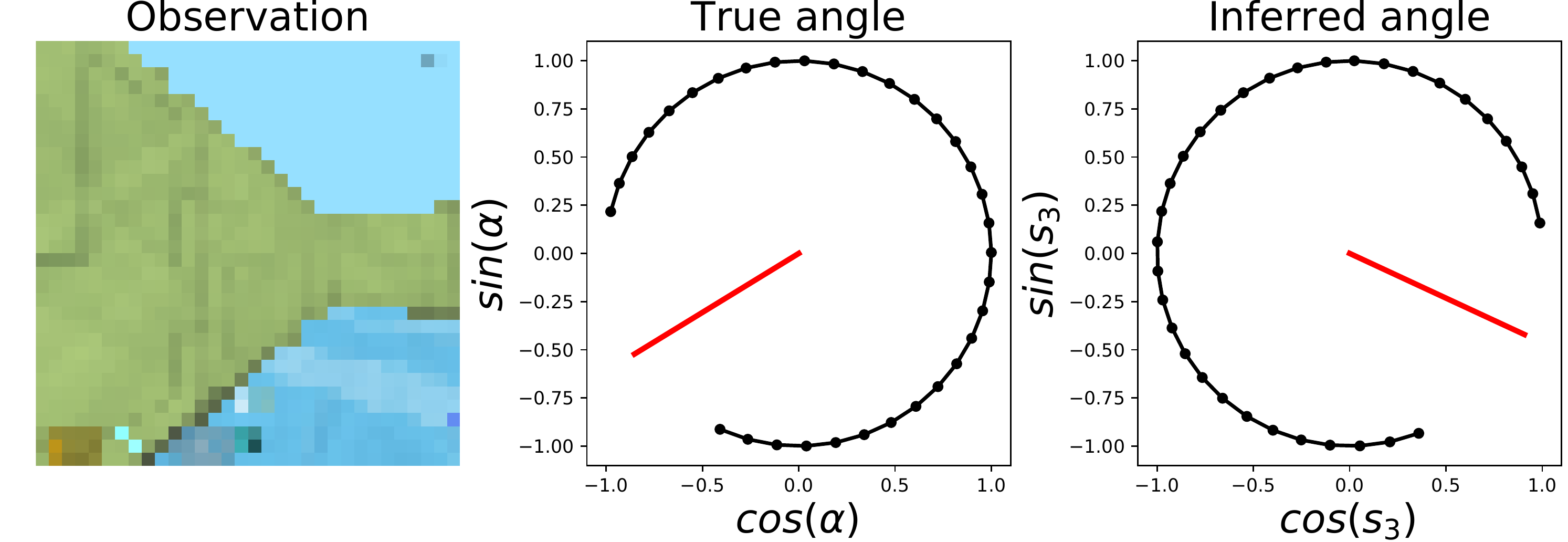}
\caption{\textbf{Rotating agent}, environment and angle information for $t=38$. The red line denotes the current orientation of the agent, the black points indicate the orientations explored in the memorization phase. $\alpha$ denotes the ground truth orientation.}\label{fig:labrot_states}
\end{subfigure}
\hfill
\begin{subfigure}[t]{0.49\textwidth}
\centering	  
\includegraphics[width=0.9\textwidth]{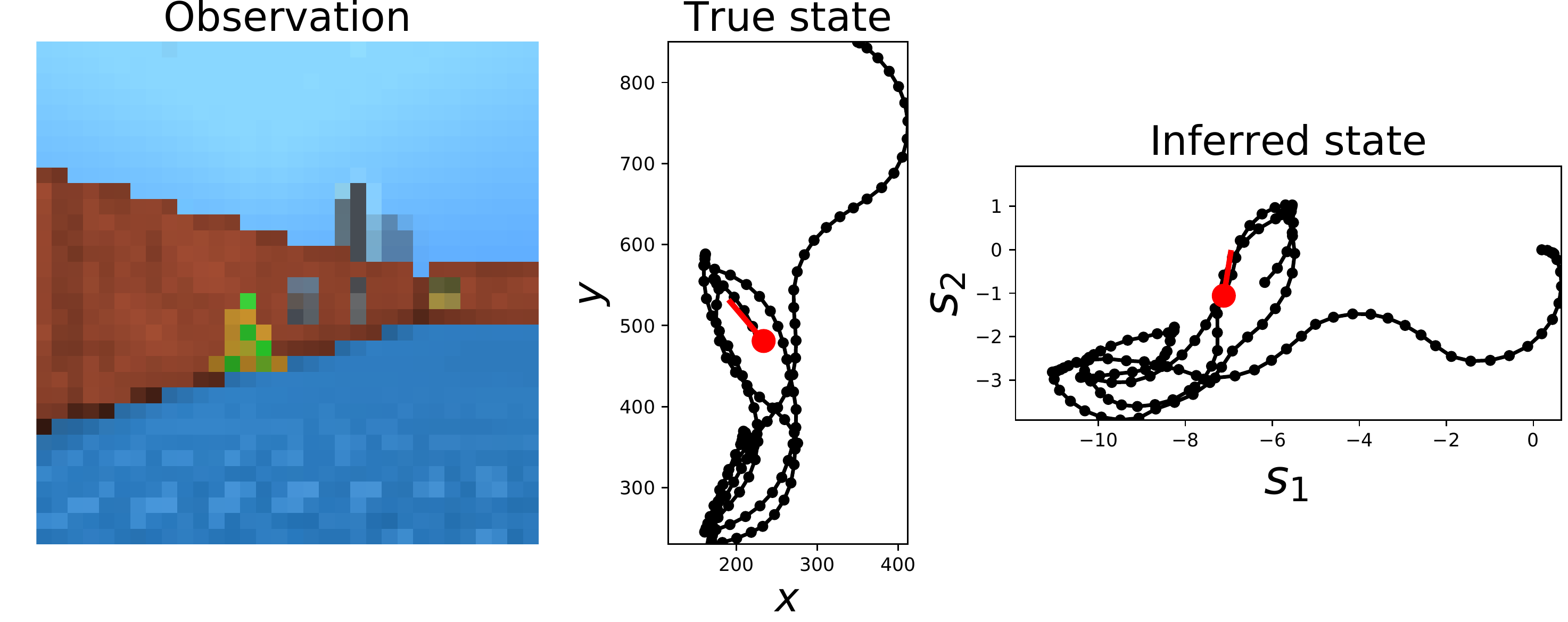}
\caption{\textbf{Walking agent}, environment and state information for $t=297$. The red circle and line represent the current position and orientation of the agent respectively, the black points indicate the states explored in the memorization phase.}\label{fig:labwalk_states}
\end{subfigure}

\vspace*{0.2cm}

\begin{subfigure}[t]{\textwidth}
\centering	  
\includegraphics[width=\textwidth]{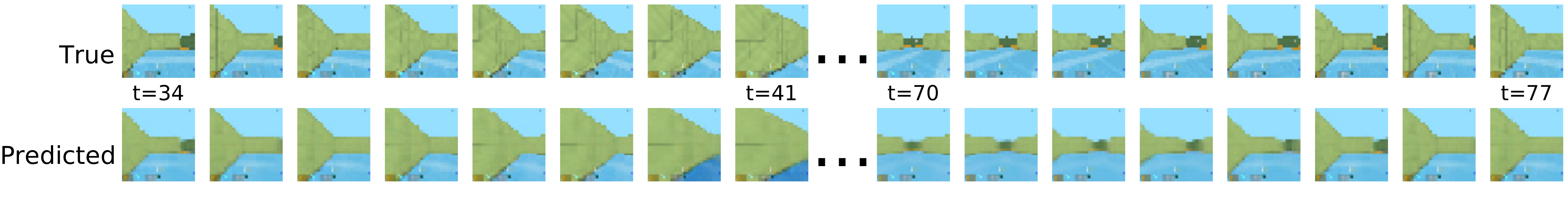}
\caption{\textbf{Rotating agent} in Labyrinth, prediction phase of a test sequence. Time steps $t_{34:41}$ and $t_{70:77}$.}\label{fig:labrot}
\end{subfigure}

\vspace*{0.1cm}

\begin{subfigure}[t]{\textwidth}
\centering	  
\includegraphics[width=\textwidth]{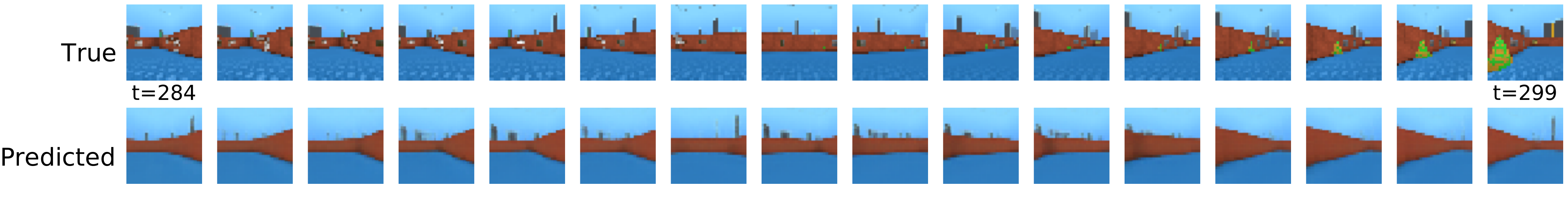}
\caption{\textbf{Walking agent} in Labyrinth, prediction phase of a test sequence. Time steps $t_{284:299}$.}\label{fig:labwalk}
\end{subfigure}
\caption{Labyrinth experiment}\label{fig:labyrinth}
\end{figure*}

We now show that the GTM-SM is able to remember the past and perform spatio-temporally coherent generations over hundreds of time steps in simulated 3D environments.
We use the \textit{Labyrinth} environment \citep{Mnih2016,Beattie2016}, procedurally-generated 3D mazes with random textures, objects and wall configurations. There are eight possible actions that can both move and rotate the agent in the maze, and we observe images from a first-person point of view.
Labyrinth can be seen as a 3D extension of the image navigation experiments in Section \ref{sec:imagenav}, but in this case the task is much harder for two main reasons that will be tackled below: (1) dealing with rotations and (2) projective transformations in a partially observable environment.

First, the state of the agent is no longer only described by the position, but also from the direction in which the agent is looking and moving. 
We need to take into account two different coordinate systems, a global one that coincides with the state-space ($\sb$-space), and one that is fixed with the agent (agent-space). In the image navigation experiments these coordinate systems coincided. The actions act in agent-space, e.g. a ``move right'' action will make the agent go right in its reference frame, but depending on its orientation this could correspond to a move to the left in $\sb$-space. 
To deal with this issues we can introduce a \textit{rotation matrix} $R$ in the state transition equation, that translates a displacement $M\ab_t$ in agent-space to a displacements $RM\ab_t$ in $\sb$-space.
More in detail, we consider a 3-dimensional state-space, and define the state transition equations as
\begin{equation}\label{eq:lab_trans}
\sb_t=\sb_{t-1} + R(\sb_{t-1}^{(3)})M \ab_t + \varepsilonb_t
\end{equation}
with $\sb_{t-1}^{(3)}$ being the 3rd component of the vector $\sb_{t-1}$ and
$$
R(\sb_{t-1}^{(3)}) = 
\begin{bmatrix}
\cos(\sb_{t-1}^{(3)}) & -\sin(\sb_{t-1}^{(3)}) & 0 \\
\sin(\sb_{t-1}^{(3)}) & \cos(\sb_{t-1}^{(3)}) & 0 \\
0 & 0 & 1 
\end{bmatrix} \ .
$$
As for the image navigation experiment, we learn the parameters of $M$. 
While we do not explicitly tell the GTM-SM to use the first two component of the state vector as a position and the third one as an angle, the model will learn to use them in this way in order to maximize the ELBO. 
In the nearest neighbor search in the DND memory we need to take into account the periodicity of the angle, e.g. that an agent oriented at $30^\circ$ or $390^\circ$ is actually looking in the same direction. 
When computing distances, instead of using $\sb_{t}^{(3)}$ we then use $\cos(\sb_{t}^{(3)})$ and $\sin(\sb_{t}^{(3)})$, that are possibly passed together with the first two components of $\sb_{t}$ through a linear layer that maps the resulting vector in a learned space where we use the Euclidean distance.

The second challenge arises from the fact that, unlike the image navigation experiment where there were a limited a number of possible positions for the agent, in the 3D labyrinth environment it is not reasonable to assume that during the memorization phase the agent will pass in all positions and look from them in all directions.
To deal with this issue, we use as VAE prior $p_\theta(\zb_t|\sb_t,\mb)$ a \textit{Generative Query Network} \citep{Eslami2018}, a neural architecture that given the frames from the closest positions retrieved from the DND memory learns to combine the different views taking into account projective transformations

\subsubsection{Rotating agent in Labyrinth}\label{sec:labrot}
In the first experiment we test the abilities of the GTM-SM to learn to model rotations with the transition model in \eqref{eq:lab_trans} as well as to combine the information from different views.
We use videos with action data of an agent that does two complete rotations while standing still in the same position. The rotational period is around $41$ time steps, but we only store in memory the first $\tau=33$ (approximately $300^\circ$). We then ask the model to generate the remaining $60^\circ$ to finish the first rotation and a whole new rotation.
From Figure \ref{fig:labrot_states} we see an example of an observation from the test data and that the model has correctly learned to use the third component of the state vector $\sb_t$ to represent the orientation of the agent. In the prediction phase in Figure \ref{fig:labrot}, we notice that the predictions from the model are very close to the ground truth, meaning that the model has learned to use the memory correctly. In particular, despite the fact that the frames from $t=33$ to $t=41$ were never seen during the memorization phase, the GTM-SM has learned to combine the information from other views.
Notice that this experiment can be seen as a more challenging version of the Labyrinth rotation experiment of \citep{Gemici2017}, that used a fully observed first rotation with a rotational period of 15 time steps.

\subsubsection{Walking agent in Labyrinth}\label{sec:labwalk}

We now use videos of a pre-trained RL agent walking in a room and solving a scavenger hunt task. 
In this case it is fundamental to extend the transition equation in \eqref{eq:lab_trans} to model more carefully the physics of the walking agent, that make the displacement at a given time step depend not only on the current action, but also on the displacement at the previous time step.
The agent is subject to momentum and friction, so that if it is moving in a certain direction at time $t-1$, it will still continue to move a bit in the same direction even at time $t$, regardless of the action $\ab_t$.
Also, despite the momentum, the displacement of the agent cannot increase indefinitely, i.e. there is saturation.
We can model this by extending the way the displacement $\db_t=M \ab_t$ is calculated in \eqref{eq:lab_trans}. 
To take into account momentum and friction, we first add to $\db_t=M \ab_t$ a damped version of the displacement at the previous time step, i.e. $\sigma(\cb_f)\odot\db_{t-1}$, where $\sigma(\cb_f)$ is a learned vector between 0 and 1 of the same size of $\db_t$ ($\sigma$ is the sigmoid function and $\odot$ represents the element-wise product).
To deal with saturation, we then limit the range of the displacements by squashing them through a $\tanh$ non-linearity that is pre-multiplied by a learned vector $\cb_s$. The resulting transition model then becomes

$
\sb_t=\sb_{t-1} + R(\sb_{t-1}^{(3)})\underbrace{\cb_s \odot \text{tanh}(\sigma(\cb_f)\odot\db_{t-1} + M \ab_t)}_{\db_t} + \varepsilonb_t
$
with $M$, $\cb_s$ and $\cb_f$ parameters to be learned. 
Notice that modelling the displacements in this way essentially corresponds to extending the state space with an additional vector $\db_t$ that models the velocities of the agent. We can then model accelerations/decelerations of the agent by non-linearly changing the velocities over time in this extended state-space.
This is a very challenging experiment, as from only one-hot encoded actions and the frames of the video we need to learn a complex transition model. Moreover, due to the low resolution of the images (32x32), small variations in state-space may be impossible to infer using only the images. To solve this, we make the reasonable assumption that at training time the agent feels its movement, i.e. the displacements. We then add a regression loss as an extra term to the objective function, that helps the GTM-SM to learn transition parameters such that the estimated $\db_t$ is close to its true value. We only add the true displacements information as a target in the loss, and never pass it directly into the model.

We let the agent walk around the room while adding information in the DND memory for $\tau=150$ time steps, and then predict during testing the following 150 time steps ($T=300$).
In Figure \ref{fig:labwalk_states}, we notice that the GTM-SM is able to learn a very accurate transition model, that provides a sufficiently good approximation of the true state even after $t=297$ time steps.
In Figure \ref{fig:labwalk} we can appreciate the memorization and long-term generation capabilities of the GTM-SM by looking at the comparison between the true and predicted frames of the video in the end of the prediction phase ($t_{284:299}$). 
We also notice in the predicted frames, that the model correctly draws the walls and the floors but fails to render the objects, probably due to difficulties in modelling with the VAE the very diverse and complex textures that form objects in this environment.
We also tested the same trained model on longer videos of larger environments with multiple rooms ($\tau=150$ and $T=450$).
As explained in detail in Appendix \ref{sec:exp_multi}, the model is able to correctly predict the textures of the environment even after 300 time steps.
\section{Related work}\label{sec:related}

A number of recent works have augmented deep generative models with learned external memories, both in the static setting \citep{Li2016,Bornschein2017} and in the temporal one \citep{Gemici2017}. 
More in general, neural networks have been combined with different memories in a wide range of tasks such as supervised learning \citep{Graves2014,Graves2016}, reinforcement learning \citep{Oh2016,Pritzel2017,Parisotto2018}, one-shot learning \citep{Santoro2016}, question answering and language modelling \citep{Sukhbaatar2015,Miller2016}. 
Each memory architecture uses different addressing mechanisms to write or read information, that are usually chosen depending on the specific application being considered.
As discussed in the introduction, our work is closely related to \citep{Gemici2017}, but more suitable for long-term generation for this task and more scalable thanks to the usage of the spatial memory architecture that exploits knowledge on the dynamics of the agent and does not require to learn a parametric memory addressing scheme.

In the deep reinforcement learning community, several works have exploited different memory architectures to store long term information to be used within an agent's policy, such as in \citep{Zaremba2015,Oh2016}. 
In particular, in \citep{Gupta2017,Gupta2017b,Zhang2017,Parisotto2018} the memory architectures have a fixed number of slots that are spatially structured as a 2D grid, and can therefore store information on the moving agent. 
Similarly to the GTM-SM, these memories are built to exploit the spatial structure of the problem, although for the different task of constructing agents that can learn to navigate and explore the environment, as opposed to the focus on generative modelling of this paper.
Simultaneous Localization And Mapping (SLAM) \citep{Smith1987,Leonard1991} is a popular technique used in robotics to estimate the position of a robot and the map of the environment at the same time using sensor data, recently applied to deep reinforcement learning for example in \citep{Bhatti2016,Zhang2017}.
It is reminiscent to the memorization phase of the GTM-SM, that could be therefore extended using ideas introduced in the visual SLAM community \citep{Taketomi2017}.

\section{Conclusion}\label{sec:conclusion}
In this work we introduced an action-conditioned generative model that uses a scalable non-parametric memory to store spatial and visual information.
Our experiments on simulated 2D and 3D environments show that the model is able to coherently memorize and perform long-term generation.
To our knowledge this is the first published work that builds a generative model for agents walking in an environment that, thanks to the separation of the dynamics and visual information in the DND memory, can coherently generate for hundreds of time steps in a scalable way.
Future work will focus on exploiting these capabilities in model-based planning by integrating the GTM-SM within an RL agent.

\section*{Acknowledgements}
We would like to thank Charles Blundell and Timothy Lillicrap for many helpful discussions, and Ivo Danihelka and Joshua Abramson for generating the datasets used in the Labyrinth experiments.

\bibliography{bibliography_gtm_dnd}
\bibliographystyle{icml2018}

\newpage
\appendix
\section{Experimental details}\label{sec:exp_details}

The models used in all experiments are implemented in Tensorflow \citep{tensorflow2015} and use the Adam optimizer \citep{kingma2014adam}.

\subsection{Image navigation experiment}\label{app:imagenav}
The training and test data sets are procedurally generated by sampling a random trajectory in randomly chosen images from the CelebA data set. The actions at each time steps are one-hot encoded (vector of size 5). The memorization phase is 256 time steps (we experimentally determined that this suffices to ensure that the agent usually reaches most positions in the environment), while the prediction one has 32 time steps during training and 256 during testing.

The VAE prior used in this experiment is obtained by creating a mixture distribution from the sufficient statistics of the frame encodings retrieved from the DND memory, whose weights are inversely proportional to the squared distances $d^{(k)}$ between $\sb_t$ and the retrieved elements $\sb_k^{(k)}$:
$$
p_\theta(\zb_t|\sb_t,\mb) =\sum_{k=1}^K w_k\Ncal(\zb_t|\zb^{(k)}) \ ; \quad w_k \propto\frac{1}{{d^{(k)}}^2+\delta} 
$$
$\delta=10^{-4}$ is added for numerical stability \citep{Pritzel2017}.
In the DND memory we store sufficient statistics, but in this experiment we use the Euclidean distance between means in the nearest-neighbor search. (Alternatively, we could use the KL divergence between the distributions).

The VAE decoder and encoder use a 3-layered convolutional architecture to parameterize mean and variance of 16-dimensional latent states, but we noticed in practice that for this experiment even standard fully connected architectures perform well. In the transition model the standard deviation of the model is $r=10^{-3}$. In the DND we retrieve the 5 nearest neighbour and use Euclidean distances between means.

The initial learning rate is $10^{-3}$, and we anneal it linearly to $5\cdot10^{-5}$ during the first 50000 updates.

\subsection{Labyrinth experiments}\label{sec:app_exp_lab}
The data sets used for the labyrinth experiments contain 120000 action-conditioned videos, of which we use 100000 for training and 20000 for testing.
Each video for the rotating agent experiments contains 80 frames. To form a training sequence we select randomly 49 consecutive frames of a video, that we split in 33 frames for the memorization phase and 16 for the prediction one. 
During testing, the prediction phase has 45 time steps.
For the walking agent experiment the videos are 300 time steps. 
Similarly to the rotation experiment, to form a training sequence we get consecutive sequences of 150+32 time steps (memorization and prediction phase respectively). 
During testing, we use 150 frames for the prediction phase.

The transition noise of the SSM has standard deviation $r=10^{-2}$. The pre-trained agent does not hit walls, therefore we do not need to handle non-linearities as in \eqref{eq:nonlin_tr}. To compute distances in the DND we map the state vectors to
$$
\widetilde{\sb}_t= 
\begin{bmatrix}
\sb_{t}^{(1)} \\
\sb_{t}^{(2)} \\
\cos(\sb_{t}^{(3)}) \\
\sin(\sb_{t}^{(3)}) 
\end{bmatrix} \ ,
$$
and optionally pass the resulting vector through a linear layer. This gives a 5-dimensional vector in a learned manifold in which we use the Euclidean distance (in our experiments, the model performed well even without the linear layer). In the DND we retrieve the 4 nearest neighbours.

In the VAE, we use convolutional encoder and decoder, and 64-dimensional latent state.
The VAE prior $p_\theta(\zb_t|\sb_t,\mb)$ used in the Labyrinth experiments is slightly more involved than the mixture prior used in the image navigation one. It is formed using a \textit{Generative Query Network} \citep{Eslami2018}, that first maps the data retrieved from memory $\{ \sb_i, \zb_i \}$ with a MLP to an embedding vector $h_t$, and then combines the embedding $h_t$ with the current state $\sb_t$, mapping the result to the mean and variance of the Gaussian prior 
\begin{align}
h_t &= f(\{ \sb_i, \zb_i \}) \nonumber\\
p_\theta(\zb_t|\sb_t,\mb) &= \Ncal( \mub(h_t, \sb_t), \sigmab(h_t, \sb_t) ).\nonumber
\end{align}

The initial learning rate is set to $3\cdot10^{-3}$, and we linearly anneal it to $5\cdot10^{-5}$ during the first 100000 updates.

\section{Videos of long-term generation}\label{sec:app_video}
Videos of long-term generation from the GTM-SM for all the experiments of this paper are available at this \href{https://drive.google.com/open?id=1WLiyLRDUIMuOWtJEDIUxrTQCOOykeCE0}{Google Drive} link (\href{https://goo.gl/RXQPTL}{goo.gl/RXQPTL}).
The videos are subdivided in folders:
\begin{enumerate}
\item \texttt{videos/image\_navigation/} contains the videos for the experiments in Section \ref{sec:imagenav}.
\item \texttt{videos/labyrinth\_rotation/} contains the videos for the experiments in Section \ref{sec:labrot}.
\item \texttt{videos/labyrinth\_walk/} contains the videos for the experiments in Section \ref{sec:labwalk}.
\item \texttt{videos/labyrinth\_walk\_multirooms/} contains the videos for the experiments in Appendix \ref{sec:exp_multi}.
\end{enumerate}
In all folders, the first video corresponds to the test sequence used to produce the figures in the paper.

\section{Walking agent in Labyrinth (multiple rooms)}\label{sec:exp_multi}

\begin{figure*}[t]
\centering
\includegraphics[width=\textwidth]{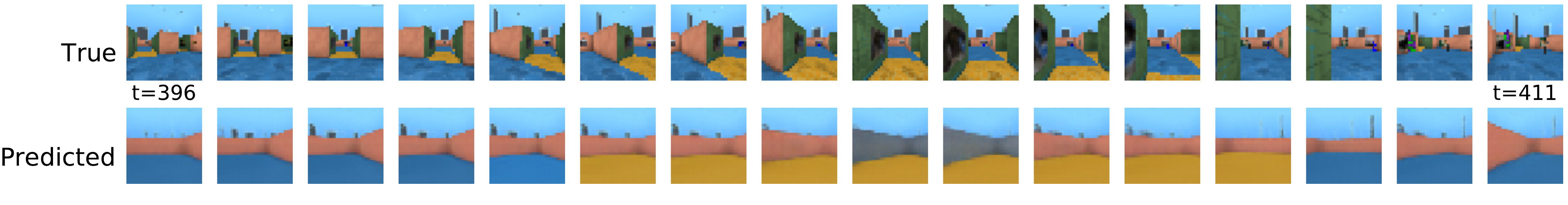}
\caption{Prediction phase for a walking agent trained on one room and tested on multiple rooms. Time steps $t_{396:411}$.}\label{fig:labwalk_multi}
\end{figure*}

We consider the same trained model used for the results in section \ref{sec:labwalk}. In section \ref{sec:labwalk}, this model was tested on videos of length $T=300$ of an agent walking in a single room, with both memorization and prediction phase of 150 time steps.
To asses the long-term memorization and localization capabilities of the GTM-SM, we now test it on videos of the same agent walking in larger environments with multiple rooms. Each video is $T=450$ time steps; we store 150 time steps in memory and we predict for 300 more.
As the model is trained on single rooms, we cannot expect the VAE to correctly generate the corridors between rooms, but we can expect the model to be able to know its position and the textures in the room (i.e. the color of the walls and of the floor).

In Figure \ref{fig:labwalk_multi} we show the predictions from the model after more than 250 time steps from the end of the memorization phase. As expected, the model fails in drawing the walls that form the corridor between the two rooms. However, we see that the GTM-SM correctly remembers the texture of rooms that it has previously visited and is able to predict the change in the color of the floor in the corridor. 
This is better viewed looking at the videos of this experiment, available in the folder \texttt{videos/labyrinth\_walk\_multirooms/} in the supplementary material.

\section{Inference network using landmark information}\label{sec:inf_landmark}
\begin{figure}[t]
\centering	  
  \begin{tikzpicture}[bend angle=45,>=latex,font=\small,scale=1, every node/.style={transform shape}]%

	\tikzstyle{obs} = [ circle, thick, draw = black!100, fill = blue!10, minimum size = 0.8cm, inner sep = 0pt]
	\tikzstyle{lat} = [ circle, thick, draw = black!100, fill = red!0, minimum size =  0.8cm, inner sep = 0pt]
	\tikzstyle{par} = [ circle, thin, draw, fill = black!100, minimum size = 0.8, inner sep = 0pt]
	\tikzstyle{det} = [ diamond, thick, draw = black!100, fill = red!0, minimum size = 1cm, inner sep = 0pt]
	\tikzstyle{inv} = [ circle, thin, draw=white!100, fill = white!100, minimum size = 0.8cm, inner sep = 0pt]
	\tikzstyle{annotation} = [rectangle, thin, draw=white, fill=white ]
	\tikzstyle{every label} = [black!100]%
	\begin{scope}[node distance = 1.4cm and 1.4cm, rounded corners=4pt]

	    \node (z_tm2) [] {};
		\node [lat] (z_tm1) [ right of = z_tm2]   {$\sb_{t-1}$};
		\node [lat] (z_t) [ right of = z_tm1] {$\sb_{t}$};
		\node (z_tp1) [ right of = z_t]{};

        \draw[post] (z_tm2) edge  (z_tm1);
		\draw[post] (z_tm1) edge  (z_t);
		\draw[post] (z_t) edge  (z_tp1);

		\node [rectangle, draw=blue] (m_tm1) [ below = 0.35cm of z_tm1]   {Memory};
		\node [rectangle, draw=blue] (m_t) [ below = 0.35cm of z_t] {Memory};

		\draw[post,draw=blue,fill=blue] (m_tm1) edge  (z_tm1);
		\draw[post,draw=blue,fill=blue] (m_t) edge  (z_t);

		\node [lat] (a_tm1) [ below = 0.15cm of m_tm1]   {$\zb_{t-1}$};
		\node [lat] (a_t) [ below = 0.15cm of m_t] {$\zb_{t}$};

		\draw[-,draw=blue,fill=blue] (a_tm1) edge  (m_tm1);
		\draw[-,draw=blue,fill=blue] (a_t) edge  (m_t);


		\node [obs] (x_tm1) [ below of = a_tm1]   {$\xb_{t-1}$};
		\node [obs] (x_t) [ below of = a_t] {$\xb_{t}$};

		\node [obs] (u_tm1) [ above of = z_tm1]   {$\ab_{t-1}$};
		\node [obs] (u_t) [ above of = z_t] {$\ab_{t}$};

		\draw[post] (u_tm1) edge  (z_tm1);
		\draw[post] (u_t) edge  (z_t);

		\draw[post] (x_tm1) edge[bend left]  (a_tm1);
		\draw[post] (x_t) edge[bend left]  (a_t);

	\end{scope}%
\end{tikzpicture}%
\caption{Inference network for the GTM-SM using landmark information. Blue arrows represent dependencies on the reversed DND memory.}%
\label{fig:ssmm_inf_landmark}%
\end{figure}
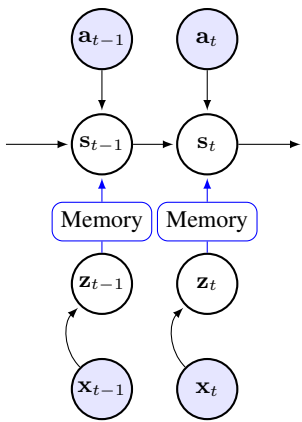

We now introduce an alternative inference network that uses the information in the DND memory to improve inference in cases in which the SSM transition model is not powerful enough to infer the correct position of the agent.
We factorize the variational approximation $q_\phi(\zb, \sb| \xb, \ab, \vb)$ as
\begin{align*}
q_\phi(\zb, \sb| \xb, \ab, \vb) & = q_\phi(\zb| \xb) q_\phi(\sb| \zb, \ab, \vb)\\
& = \prod_{t=\tau+1}^T q_\phi(\zb_t| \xb_t) q_\phi(\sb_t| \sb_{t-1}, \ab_t, \zb_t, \mb) \ .
\end{align*}
A graphical representation of the inference network of the GTM-SM is shown in Figure \ref{fig:ssmm_inf_landmark}.
$q_\phi(\zb_t| \xb_t)$ is an inference network that outputs the mean and variance of a Gaussian distribution, as typically done in VAEs. 
The structured variational approximation $q_\phi(\sb_t| \sb_{t-1}, \ab_t, \zb_t, \mb)$ retains the temporal dependency among the state variables and exploits the information stored in the memory $\mb$.
We define this approximation to depend on:
\begin{enumerate}
\item The \textit{prior belief} $p_\theta(\sb_{t}|\sb_{t-1}, \ab_{t})$. If at time $t-1$ we were at a given position, this dependence captures the fact that at time $t$ we cannot be too far from it. This is the same dependence we used in Section \ref{sec:inf}.
\item \textit{Landmark information}, obtained by querying the DND memory in the reverse direction with respect to the VAE prior, i.e., considering the frame encodings $\zb_i$ as keys and the states $\sb_i$ as values. 
At each time step the agent can then check whether it has already seen the current frame encoding $\zb_t$ in the past, and exploit this information when computing the inferred position of the agent. 
We use $\zb_t$ to query the reversed-DND, retrieving triplets $\{(\delta^{(k)}, \zb^{(k)},\sb^{(k)}), k=1, .. ,K^\prime\}$ that are used in the computation of the parameters of $q_\phi(\sb_t| \sb_{t-1}, \ab_t, \zb_t, \mb)$. Here, $\delta_i^{(k)}$ represents a distance in $\zb$-space.
\end{enumerate}

We define $q_\phi(\sb_t| \sb_{t-1}, \ab_t, \zb_t, \mb)$ to be a Gaussian density, whose mean $\mub_q$ and variance $\sigmab^2_q$ are the outputs of a neural network that merges the sufficient statistics $\mub_p$ and $\sigmab^2_p$ of the prior $p_\theta(\sb_{t}|\sb_{t-1}, \ab_{t})$, and the ones of the states $\sb_i^{(k)}$ retrieved using landmark information: $\mub_k, \sigmab^2_k, k=1:K^\prime$. We assume that we stored in the DND the mean and the variance of Gaussian latent states. The posterior mean is obtained as 
$$
\mub_q=\mub_p+\sum_{k=1}^{K^\prime} \beta_k(\mub_k-\mub_p) \ ,
$$
where $\beta_k\in[0, 1]$ is the output of a simple neural network with input $\delta_i^{(k)}$.
The inference network can then learn to assign a high value to $\beta_k$ whenever the distance in $\zb$-space is small (i.e. the current observation is similar to a frame stored in the DND), so that the prior mean is moved in the direction of $\mub_k$. Similarly, the posterior variance can be computed starting from the prior variance using another neural network: $\log \sigmab^2_q = \log \sigmab^2_p + NN(\delta_i^{1:{K^\prime}},\sigmab^2_{1:{K^\prime}}, \sigmab^2_p)$.

With this choice for the inference network, the ELBO of the GTM-SM becomes
\begin{align*}
 \Fcal&(\theta , \phi)  =\sum_{t=\tau+1}^T \Ebb_{q_\phi(\zb_t|\xb_t)} \left[ \log p_\theta(\xb_t|\zb_t) \right] +\\ 
 &  - \Ebb_{q_\phi^\star(\sb_{t})} \left[ \mathit{KL} [q_\phi(\zb_t|\xb_t) ||  p_\theta(\zb_t|\sb_t,\mb) ]  \right] +\\
 & - \Ebb_{q_\phi^\star(\sb_{t-1})} \left[ \mathit{KL}\left[ q_\phi(\sb_t|\sb_{t-1},\ab_t, \zb_t, \mb) || p_\theta(\sb_t|\sb_{t-1},\ab_t)  \right]  \right] \ .
\end{align*}
with
$$
q_\phi^\star(\sb_{t}) = \int q_\phi(\sb_{t}|\sb_{t-1},\ab_{t}, \zb_{t}, \mb_{1:t-1}) q_\phi^\star(\sb_{t-1}) \ \drm \sb_{t-1}\ .
$$
Notice in particular the additional KL term for the SSM.

\subsection{Image navigation with obstacles}
\begin{figure}[t]
\centering
\includegraphics[width=0.48\textwidth]{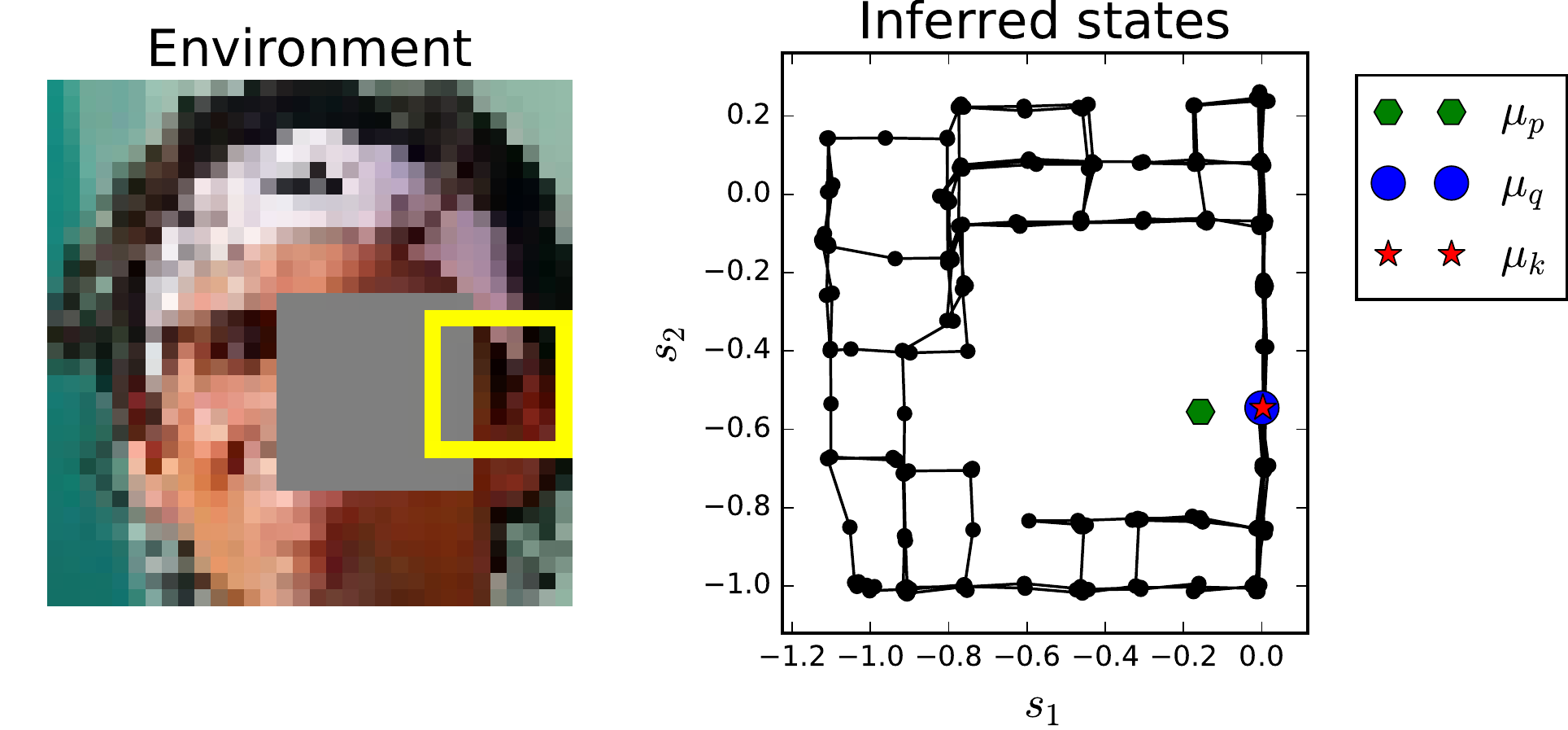}
\caption{Image navigation experiment with obstacles. The agent (yellow square) cannot cross the ostacle (the gray squared area).}\label{fig:obstacle_nav_merge}
\end{figure}

We extend the image navigation experiments of Section \ref{sec:imagenav} adding obstacles to the environment as illustrated in Figure \ref{fig:obstacle_nav_merge} (left). 
We use displacement information as in the Labyrinth experiment of Section \ref{sec:labwalk}.
The obstacles appear in random positions in each sequence, therefore we cannot learn a prior transition model that captures these non-linear dynamics. 
However, when doing inference the model can use its knowledge on the current frame $\xb_t$ (that is not available during the prediction phase) to infer its position by exploiting landmark information.

To illustrate this we can look at the example in Figure \ref{fig:obstacle_nav_merge} (right).
At time $t-1$, the position $\sb_{t-1}$ of the agent coincides with the red star. The agent's position together with the corresponding observation $\xb_{t-1}$ (the yellow square) will be inserted in the DND memory. 
At time $t$, the agent receives a ``move left'' action; the prior transition probabilities will then predict that the agent has to move to the left (the green hexagon).
Due to the presence of the obstacle however, the agent does not move, meaning that $\xb_{t}$ will be the same as $\xb_{t-1}$.
Querying the DND in the reverse direction the model will then know that the inferred state (the blue dot) should be the same as the position at the previous time step that was stored in the DND (the red star).

\end{document}